\documentclass{article}

\usepackage[preprint]{template/corl_2026} 
\usepackage{graphicx}
\usepackage{subcaption}
\usepackage{wrapfig}
\usepackage{float}
\usepackage{flafter}
\usepackage{amsmath}
\usepackage{amssymb}
\usepackage{tabularx}
\title{ViBe: Visual Behavior Adaptation for Perceptive Humanoid Whole-Body Control}

\author{
  Lokesh Krishna,\,  Sarvesh Venkatesan,\, An Zhang,\, Quan Nguyen \\
  University of Southern California \\
  project page: \href{https://lok-i.github.io/vibe-control/}{\texttt{lok-i.github.io/vibe-control}}
}

\begin{document}
\maketitle
\vspace{-1.5em}


\begin{figure}[H]
\centering
\includegraphics[width=\linewidth]{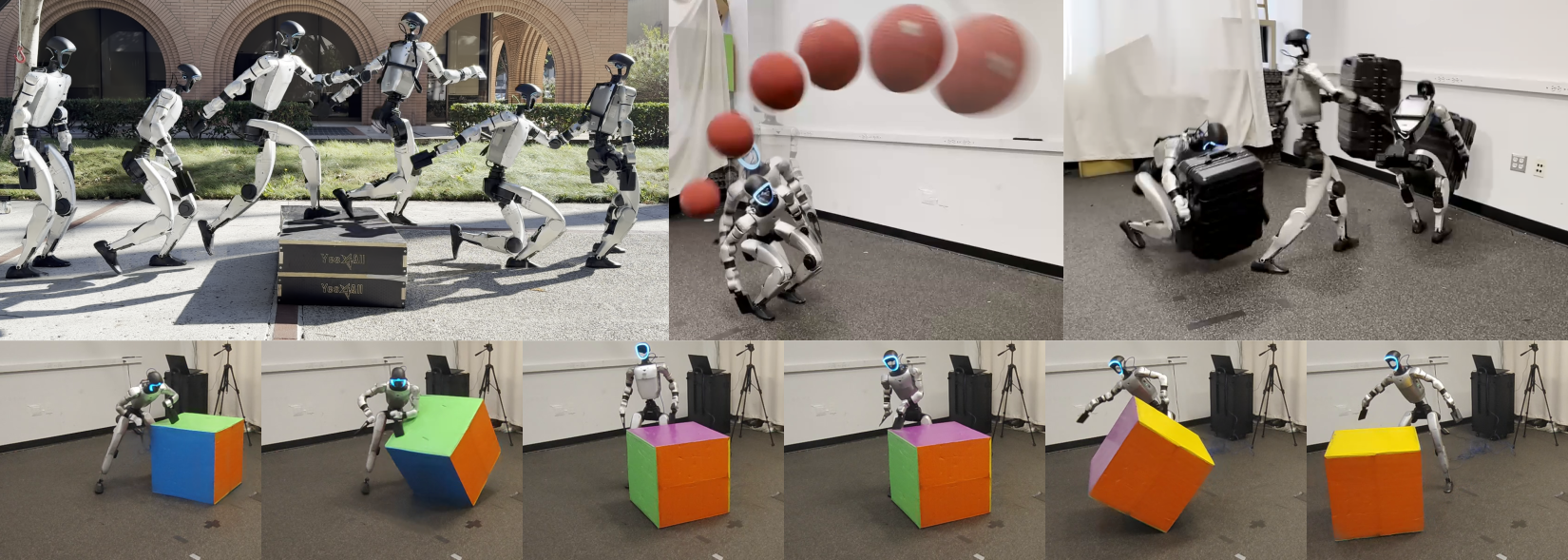}
\caption{\textbf{Visual behavior adaptation.} A single post-training recipe adapts a pre-trained
humanoid tracker to perceptive parkour, dodgeball, loco-manipulation and cube reorientation.}
\label{fig:overview}
\end{figure}

\begin{abstract}
Motion tracking provides a scalable recipe for humanoid whole-body control. By design, the
resulting trackers lack exteroceptive feedback hence reacting to the environment remains
the responsibility of a higher-level planner. Existing perceptive controllers train
geometry-only encoders from scratch, trading semantics for sim-to-real ease, and typically
rely on teacher-student distillation for a task of interest. We present ViBe, a post-training
framework for adapting motion trackers to perceptive control tasks. We leverage pre-trained
visual encoders with a multi-query extractor module to learn task-relevant perceptive
feedback. This feedback is grafted onto the tracker's input via low-rank adapters, enabling
parameter-efficient fine-tuning. Given a task reward and a reference dataset, this modular
controller can be adapted directly via policy optimization. Across four tasks, ViBe shows
zero-shot sim-to-real transfer spanning perceptive walking on curbs and parkour, 
Repose Cube, omni-object loco-manipulation, and dodgeball, with visually
robust performance across outdoor, low-light, and RGB distractor conditions. Finally, we
solve a goal-oriented Repose Cube task with a deliberately simple planner, demonstrating
the efficacy of perceptive controllers, adapted by our approach. 
\end{abstract}



\section{Introduction}
\label{sec:intro}

Humanoid whole-body control has found an effective and scalable approach: learn to mimic large
collections of human motion on flat terrain. Increasing the diversity of reference motions, model capacity,
and training compute produces controllers that preserve natural motion across a broad
behavioral repertoire~\cite{doi:10.1126/scirobotics.aed4592,doi:10.1126/scirobotics.adx8924}.
These trackers are useful motor priors, but they are intentionally designed blind to their environment. 
When a tracker encounters an environment constraint or external dynamics, the tracker
has no observation to react or adapt its behavior to solve the task.

A conventional hierarchy resolves this mismatch above the controller. Perception builds an
environment representation, a planner modifies the reference, and the tracker executes it. This
separation has enabled perceptive locomotion~\cite{10138309} and interactive whole-body
control~\cite{doi:10.1126/scirobotics.aed4592}, yet it places every environment-dependent
correction at the planning level. Such an architecture limits the lower-level modules from
controlling the contact placement to reactively interact with the environment, expecting a plan 
to be sufficiently accurate. Such an architecture results in
missing local corrections and visual reflexes like correcting foot placement over curbs,
adjusting a grasp on an object, or dodging an incoming ball.

Learning-based perceptive control exposes these corrections directly to a controller. Recent
systems traverse terrain~\cite{zhang2026rpllearningrobusthumanoid,
wu2026perceptivehumanoidparkourchaining,zhuang2026deepwholebodyparkour} and interact with non-flat scenes~\cite{videomimic}, and 
learn visual loco-manipulation~\cite{he2025viralvisualsimtorealscale}. Most approaches train task-specific perceptive
encoders and trackers and rely on teacher-student
distillation~\cite{yin2025visualmimic,he2026ultra,zhang2026rpllearningrobusthumanoid,
zhuang2026deepwholebodyparkour}. On the other hand,
attention-based encoders have recently shown that robot state can select control-relevant geometric feedback
from dense observations for locomotion~\cite{doi:10.1126/scirobotics.adv3604,11563611}. However, a
generality-preserving approach that combines a pre-trained vision encoder with a pre-trained
whole-body tracker, adapting them across diverse humanoid perceptive-control tasks remains
largely unexplored.

Our approach introduces an interface between the two pre-trained components. A frozen vision
encoder produces dense tokens, from which a compact cross-attention extractor selects task-relevant
visual feedback using the task command, global CLS token, and robot proprioception.
Low-rank adapters inject this feedback into the frozen whole-body tracker, forming a visual
bypass that leaves both pre-trained components intact. Given a reference-motion dataset and a
task reward, we train the extractor and adapters through policy optimization,
without teacher-student distillation or auxiliary representation losses.

Our contributions are twofold:
\begin{itemize}
    \item a visual behavior adaptation framework that bridges a frozen vision encoder and a
    frozen whole-body tracker through a task-conditioned extractor and low-rank adapters,
    post-trained directly with reinforcement learning.
    \item zero-shot sim-to-real transfer across four tasks: perceptive locomotion, object
    reorientation, loco-manipulation, and dodgeball, with controlled ablations
\end{itemize}

\section{Related Works}
\label{sec:related}
\begin{figure}[!t]
\centering
\includegraphics[width=\linewidth]{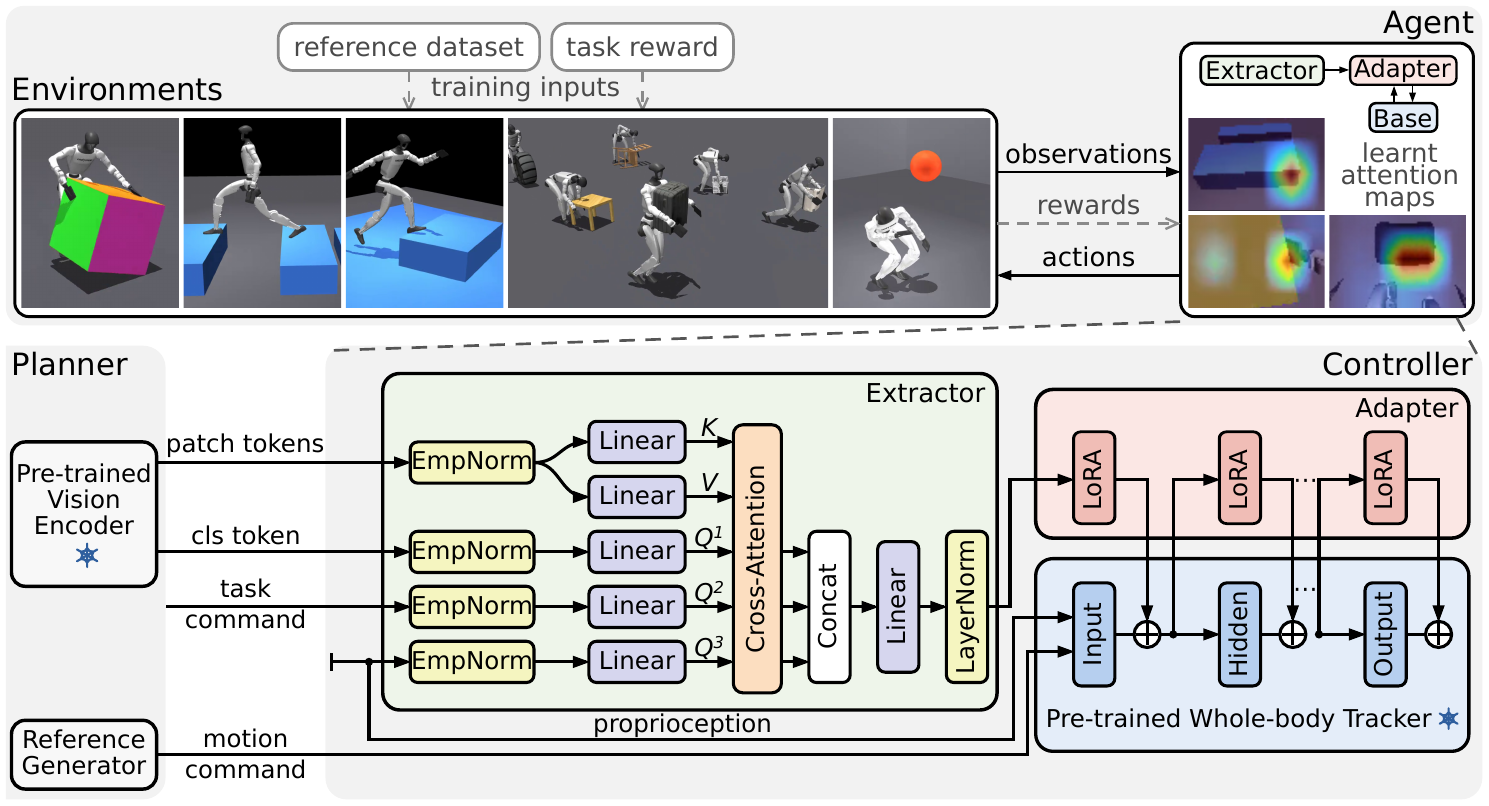}
\caption{\textbf{ViBe bridges visual and motor priors.}
\emph{Top:} Each task trains the same agent architecture from a reference-motion dataset and
task reward. \emph{Bottom:} An extractor uses global image context, task commands, and
proprioception to attend to patch tokens from a frozen vision encoder. The percept is grafted via 
LoRA adapters to the frozen whole-body tracker.}
\label{fig:system}
\end{figure}

\paragraph{Humanoid whole-body control.}
Reference tracking turns motion data into whole-body humanoid trackers
~\cite{peng2018deepmimic,doi:10.1126/scirobotics.adx8924}. Scaling motion data and model
capacity yields a broad whole-body prior ~\cite{doi:10.1126/scirobotics.aed4592,qi2026humanoidgptscalingdatastructure}.
Recent extensions add history-based adaptation for tracking under dynamic disturbances
~\cite{zhang2025trackmotionsdisturbances} or model environment-conditioned behavior for
interaction-aware control~\cite{cheng2026gigabrainwbc05behaviorworldmodel}. Our work retains the
pre-trained tracker and learns a perceptive adaptation.

\paragraph{Perceptive robot control.}
Classical systems separate perception, planning, and control through strict hierarchies and
explicit contracts~\cite{10138309}. Learning-based approaches need not follow these
architectural constraints and can learn pathways that bypass such interfaces. Existing
approaches map exteroception to actions for
quadruped locomotion~\cite{doi:10.1126/scirobotics.abk2822,pmlr-v205-agarwal23a}, humanoid
terrain traversal~\cite{zhang2026rpllearningrobusthumanoid,
wu2026perceptivehumanoidparkourchaining,zhuang2026deepwholebodyparkour,
sun2025learningperceptivehumanoidlocomotion}, dexterous
manipulation~\cite{singh2025dextrahrgbvisuomotorpoliciesgrasp}, and whole-body
interaction~\cite{videomimic,he2025viralvisualsimtorealscale}. A common sim-to-real recipe
trains an expert with privileged observations and distills it into a student with a perceptive
encoder trained from scratch. ViBe instead post-trains a deployable visual actor directly via
policy optimization.

\paragraph{Visual representations and attention.}
Pre-trained encoders provide transferable visual features without learning perception from
each task's simulated images~\cite{shang2024theia,simeoni2025dinov3,
tschannen2025siglip2multilingualvisionlanguage}. For control, attention can use robot state to
select relevant regions of a terrain map or dense depth observation
~\cite{doi:10.1126/scirobotics.adv3604,11563611}. ViBe applies the same selection principle to
visual tokens shared across perceptive whole-body control tasks. The extractor neither
predicts privileged observations nor minimizes auxiliary learning objectives but is trained
only through the downstream task objective.

\section{Approach}
\label{sec:approach}

As a post-training framework, ViBe learns only the compact extractor--adapter pathway, leaving both pre-trained priors
intact as shown in Fig.~\ref{fig:system}.

\subsection{Formulation}

We formulate each downstream task as an episodic Markov decision process with partial actor
observations. The deployable actor observation contains the robot's proprioceptive history, a future window of the reference motion,
the egocentric camera image, and an optional task command. The policy outputs actions to construct PD targets for each actuated
degree of freedom~\cite{doi:10.1126/scirobotics.aed4592}. Following~\cite{peng2018deepmimic}, the reward combines motion tracking with a task reward
like object tracking, object goal, or avoidance:

\begin{equation}
    \max_{\theta}\quad
    \mathbb{E}_{\pi_{\theta}}
    \left[\sum_{t=0}^{T}\gamma^t
    \big(r_{\mathrm{track}}+r_{\mathrm{task}}\big)\right].
    \label{eq:objective}
\end{equation}

We use an asymmetric actor--critic~\cite{pinto2018asymmetric}: privileged observations
are available to the critic, while the actor remains consistent with deployment.
We optimize the three-module agent with PPO~\cite{schulman2017proximal, schwarke2025rslrllearninglibraryrobotics}.
For the base whole-body tracker, we use SONIC~\cite{doi:10.1126/scirobotics.aed4592} and
adapt its decoder. We zero-initialize the adapters so the policy initially matches the base
tracker, and keep the policy's action standard deviation fixed throughout adaptation.
ViBe preserves the base tracker input-outputs and adds three
components: a frozen visual encoder that maps the egocentric image to visual tokens, an
extractor that selects task-relevant visual feedback, and low-rank adapters that inject this
feedback into the tracker. We train one adapted policy per task using the same architecture
and training recipe.

\begin{table}[!t]
\centering
\footnotesize
\caption{\textbf{Overview of Task Configurations.}}
\label{tab:task-design}
\renewcommand{\arraystretch}{1.25}
\renewcommand{\tabularxcolumn}[1]{m{#1}}
\begin{tabularx}{\linewidth}{|>{\raggedright\arraybackslash}m{0.17\linewidth}|>{\raggedright\arraybackslash}X|
  >{\raggedright\arraybackslash}m{0.19\linewidth}|>{\raggedright\arraybackslash}m{0.14\linewidth}|}
\hline
\textbf{Task} & \textbf{Reference, Num. Clips} & \textbf{Task Reward} &
\begin{tabular}[c]{@{}l@{}}\textbf{Task}\\\textbf{Command}\end{tabular} \\
\hline
Repose Cube & Custom in-house, 78 & Up-face (ours) & Face one-hot \\
\hline
Walk / Parkour & \begin{tabular}[c]{@{}l@{}}GRAIL~\cite{xie2026grailgeneratinghumanoidlocomanipulation}, 63\\OmniRetarget~\cite{yang2026omniretargetinteractionpreservingdatageneration}, 9\end{tabular} & Tracking~\cite{peng2018deepmimic} & Root twist \\
\hline
\begin{tabular}[c]{@{}l@{}}Omni-object\\loco-manipulation\end{tabular} & Custom in-house, 6 &
\begin{tabular}[c]{@{}l@{}}Object /\\contact~\cite{wu2026sugar}\end{tabular} & Goal pose \\
\hline
Dodgeball & Nominal pose, 1 & Clearance~\cite{mu2025smp} & --- \\
\hline
\end{tabularx}
\end{table}

\subsection{Task-based Visual Extractor}

The extractor was designed to address the question: which image patches matter to the downstream task? Our default vision encoder is a
Theia-Tiny ViT~\cite{shang2024theia}, operating
on RGB images, returning a spatial grid of
patch tokens and one global token. We use the patch tokens as keys and values for a single-head
cross-attention layer~\cite{vaswani2017attention}. Separate, normalized query projections are
formed from three signals: the global image token, robot proprioception and optionally the task command.

Each query produces an attention-weighted summary of the patch grid, as shown in
Fig.~\ref{fig:system}. We concatenate these
summaries, project them to a fixed 128-dimensional \emph{percept}, and apply layer normalization.
This fixed output size decouples the controller interface from image resolution and the encoder's latent dimensions.
Note that the query signals guide patch selection but do not enter the percept directly. While proprioception and
reference commands retain their direct paths to the controller, the task commands enter only
through visual selection. 

The \emph{percept} conditions rank-$16$ LoRA adapters in the frozen tracker's
decoder~\cite{hu2022lora}. We apply PPO updates only to the extractor and adapters, keeping
all other modules frozen and restricting adaptation to task-specific corrections of the
motion prior.

\subsection{Reference-Phase Annealing}

We observed that policies trained with uniform Reference State Initialization (RSI) can
specialize to short motion segments but fail to traverse the entire reference. Additionally,
the reference phase often tracks task progress, with later phases
corresponding to later stages of task completion~\cite{11127813}. These observations motivate reference-phase
annealing as a complementary curriculum. Let $\phi^{\max}$ denote the upper support of the normalized-phase distribution.
At policy iteration $k$, given an annealing period of $K$, we set $\phi_k^{\max}$
and sample the RSI phase $\phi_0$ as
\begin{equation}
    \phi_k^{\max}=\max\!\left(0,1-\frac{k}{K}\right),
    \qquad
    \phi_0 \sim \operatorname{Uniform}\!\left(0,\phi_k^{\max}\right).
    \label{eq:phase-annealing}
\end{equation}
At the start of training, resets span the full reference, making the curriculum identical to
uniform RSI. As training proceeds, the interval contracts toward frame zero. After $K$
updates, every episode begins at the start and thus forces mastering the full reference.

\section{Results}
\label{sec:result}

\subsection{Performance and Comparisons}

\begin{figure}[!t]
\centering
\includegraphics[width=\linewidth]{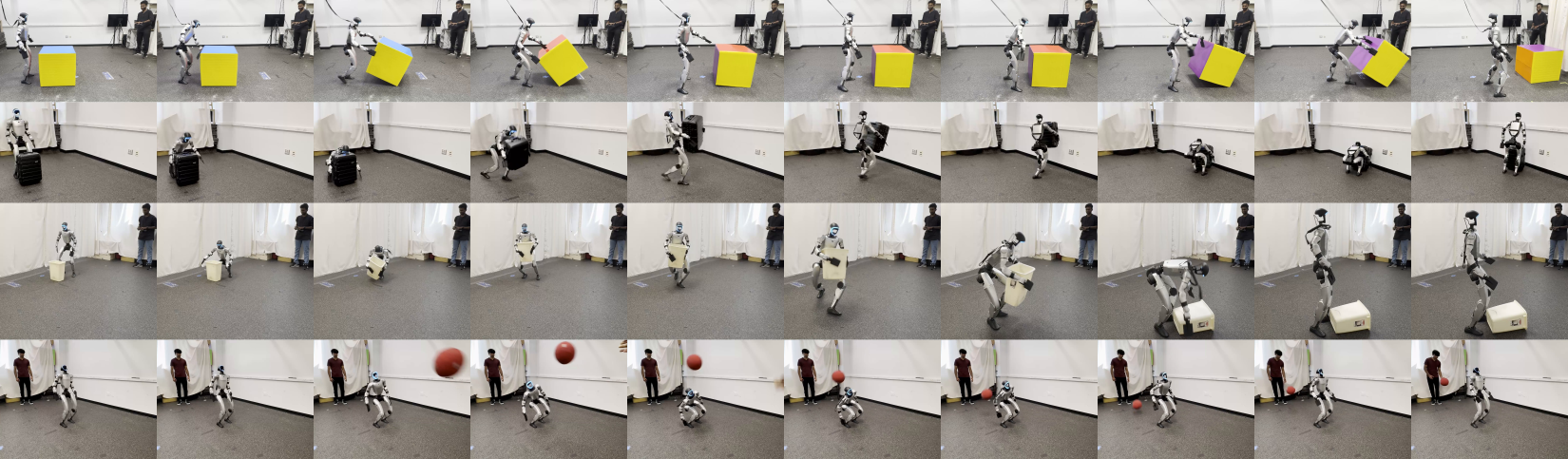}
\caption{\textbf{ViBe adapts reference motion to task dynamics.}
Zero-shot hardware rollouts show cube reorientation, suitcase transport, trash-can transport,
and dodgeball (top to bottom). Each row contains ten equally spaced frames from one
\mbox{uninterrupted fixed-camera rollout.}}
\label{fig:locomanip}
\end{figure}

\paragraph{Experimental setup.}
We evaluate on four perceptive control tasks on a Unitree G1. Table~\ref{tab:task-design} summarizes the key defining factors:
reference dataset, task-specific reward, and task command. Together, the tasks expose complementary forms of visual adaptation:

\begin{enumerate}
    \setlength{\itemsep}{0.2em}
    \setlength{\parskip}{0pt}
    \setlength{\parsep}{0pt}
    \item \textbf{Repose Cube.} Inspired by vision-based in-hand reorientation
    ~\cite{doi:10.1177/0278364919887447}, we define a whole-body variant: the humanoid tracks
    the reference while ensuring the cube tracks the object reference to result in the end-frame's face
    is on top.
    \item \textbf{Perceptive Walk and Parkour.} The two variants provide complementary
    motion-terrain pairings. GRAIL maps many walking clips per terrain, while each
    OmniRetarget parkour clip is paired with its own terrain and each task environment has 9 distinct terrains.  
    \item \textbf{Omni-object Loco-manipulation.} We place distinct objects in one environment
    and train a shared policy across them. This was to discourage memorization of a single geometry
    and makes the extractor attend to the geometry required for each interaction.
    \item \textbf{Dodgeball.} Finally, dodgeball separates task success from reference
    tracking. The reference remains a nominal standing pose, but avoiding a moving ball requires
    dynamic perceptive adaptation away from that pose.
\end{enumerate}

We implement each task in mjlab~\cite{zakka2026mjlablightweightframeworkgpuaccelerated}, built for
MuJoCo Warp~\cite{todorov2012mujoco,googledeepmind2026mujocowarp}. Thanks to its GPU batch
renderer~\cite{mustafah2026mjwarprender}, training at the necessary resolution ($112\!\times\!63$) ends within
a reasonable wall time (under two days) across different GPUs: NVIDIA RTX 3090, RTX 5090, and L40S.
Policies run at a control rate of 50~Hz, which is also the frame rate of the
head-mounted Intel RealSense D435i camera at the same resolution. Unless stated otherwise, every task
uses exactly the same architecture: a frozen Theia-Tiny encoder, the learnable extractor
and rank-16 adapters. We randomize camera extrinsics, lighting, ground appearance, robot
parameters, and external perturbations necessary for sim-to-real transfer.

\paragraph{Zero-shot transfer.}
Fig.~\ref{fig:overview} summarizes the hardware evaluation. Post adaptation, ViBe preserves
the essence of the reference motion while learning minimal desirable deviation in contact placement
and motion to solve the task. Across tasks, spanning distinct control regimes: 
the robot adjusts foot placement on raised terrain and
gains momentum for a jump, learns to
reposition itself while carrying large objects, turns a cube through repeated contacts, and
departs from a stationary reference during dodgeball (Fig.~\ref{fig:locomanip}),
separate task-specific policies trained with the same adaptation recipe.

\begin{wrapfigure}[14]{r}{0.48\textwidth}
\vspace{-2.5em}
\centering
\includegraphics[width=\linewidth]{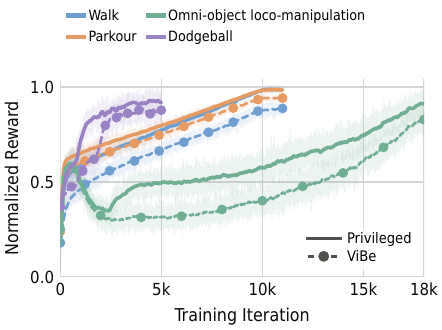}
\captionsetup{font=small}
\caption{ViBe approaches privileged-observation performance.}
\label{fig:training-curves}
\vspace{-1.0em}
\end{wrapfigure}

\begin{figure}[!t]
\centering
\begin{subfigure}[b]{0.485833\linewidth}
  \includegraphics[width=\linewidth]{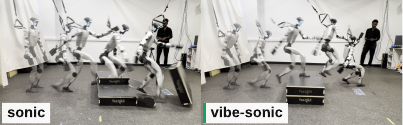}
  \caption{Perceptive Parkour}\label{fig:compare-parkour}
\end{subfigure}\hfill
\begin{subfigure}[b]{0.506619\linewidth}
  \includegraphics[width=\linewidth]{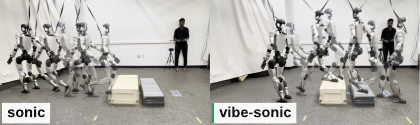}
  \caption{Perceptive Walk}\label{fig:compare-walk}
\end{subfigure}
\caption{\textbf{Visual adaptation enables terrain-aware tracking.}
With the same SONIC base and reference clips, both in (a) parkour  and (b) walk: the blind tracker (left) fails to traverse the terrain, 
whereas ViBe-SONIC (right) adjusts its motion and succeds in both behaviors.}
\label{fig:compare}
\end{figure}

\paragraph{Privileged observation vs. visual learning.} We compare direct visual learning with
privileged-observation training to validate whether post-training requires
teacher-student distillation.  Trained directly from visual feedback, ViBe reaches 89.6\%,
95.5\%, 90.3\%, and 94.7\% of the corresponding privileged policy on perceptive
walk, perceptive parkour, omni-object loco-manipulation, and dodgeball, respectively. Across
tasks, ViBe asymptotes privileged-observation performance without teacher action targets, indicating that
teacher-student distillation is unnecessary for post-training in our setting.

\paragraph{\mbox{Blind tracking vs. visual adaptation.}}\hspace{-0.67em}
The controlled comparison in Fig.~\ref{fig:compare} isolates the visual bypass. Both policies
share the same base and reference motion, but only ViBe-SONIC observes the scene. On Perceptive
Walk, the blind tracker collides with the curb, preventing ascent and causing global root drift;
ViBe-SONIC instead adjusts its foot placement and traverses the curb. On Perceptive Parkour,
ViBe-SONIC also tracks the motion velocity more closely, building the momentum needed to
complete the jump. With the base and reference fixed, these differences reflect visual
adaptation rather than tracker quality.

\begin{figure}[H]
\centering
\includegraphics[width=\linewidth]{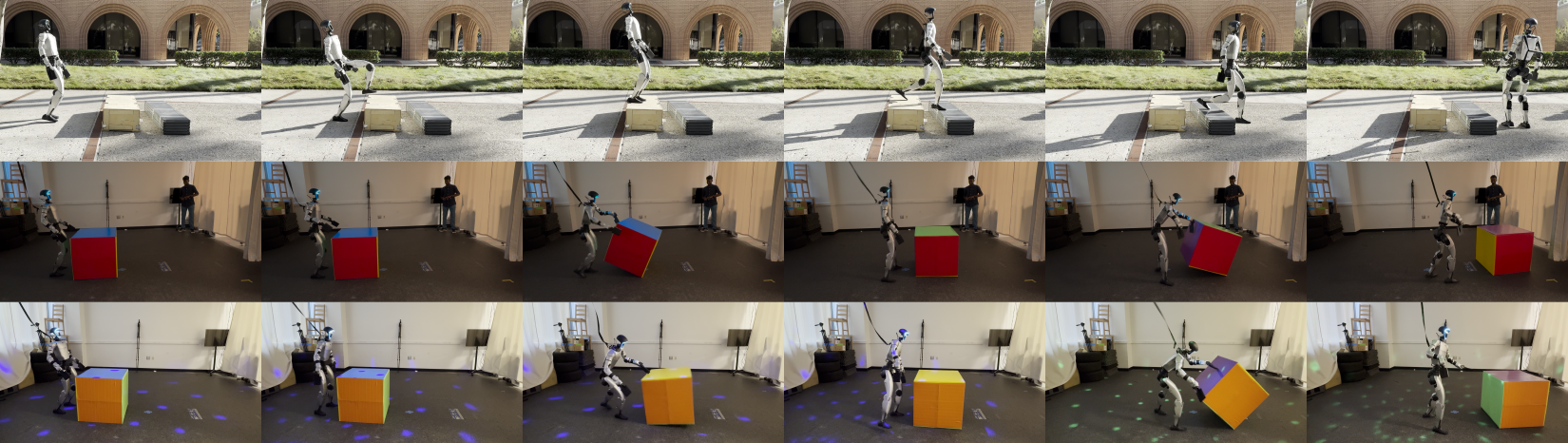}
\caption{\textbf{Adapted policies tolerate substantial appearance shifts.}
Zero-shot hardware rollouts show Perceptive Walk outdoors in direct sunlight(top), Repose Cube in
low light (center), and Repose Cube under moving colored illumination (bottom)}
\label{fig:vis-robust}
\end{figure}

\paragraph{Visual robustness.}
Policies trained with the same frozen vision backbone remain effective under out-of-distribution
visual conditions unseen during training. Fig.~\ref{fig:vis-robust} shows perceptive walk outdoors 
in bright midday sunlight and repose cube with low light and RGB distractors. These trials demonstrate
tolerance to static appearance shifts; but do not establish invariance to unmodeled dynamic physical
distractors, which remain a limitation.

\subsection{Solving Tasks with a Planner}

To demonstrate the efficacy of our perceptive control, we design a deliberately
simple planner to solve the Repose Cube task. Like Sledgehammer and Sune in Rubik's
Cube solving~\cite{speedsolvingwiki2026sledgehammer,speedsolvingwiki2026sune}, the planner repeatedly
executes a fixed routine: right flip, right flip, front flip, until the requested color is on top.
The planner has no dynamics model orforward predictive rollouts and merely chooses the clip from 
the fixed dataset. Since the adapted controller handles object localization, contact placement, and recovery from the
stance mismatches through learnt visual cues, even a simple planner can solve the task. The
hardware rollouts in Fig.~\ref{fig:repose-planner} show three goal colors with more in the supplementary
video. Thus, we demonstrate that a perceptive controller capable of local corrections can effectively
solve tasks in conjunction with a higher-level planner.

\begin{figure}[H]
\centering
\includegraphics[width=\linewidth]{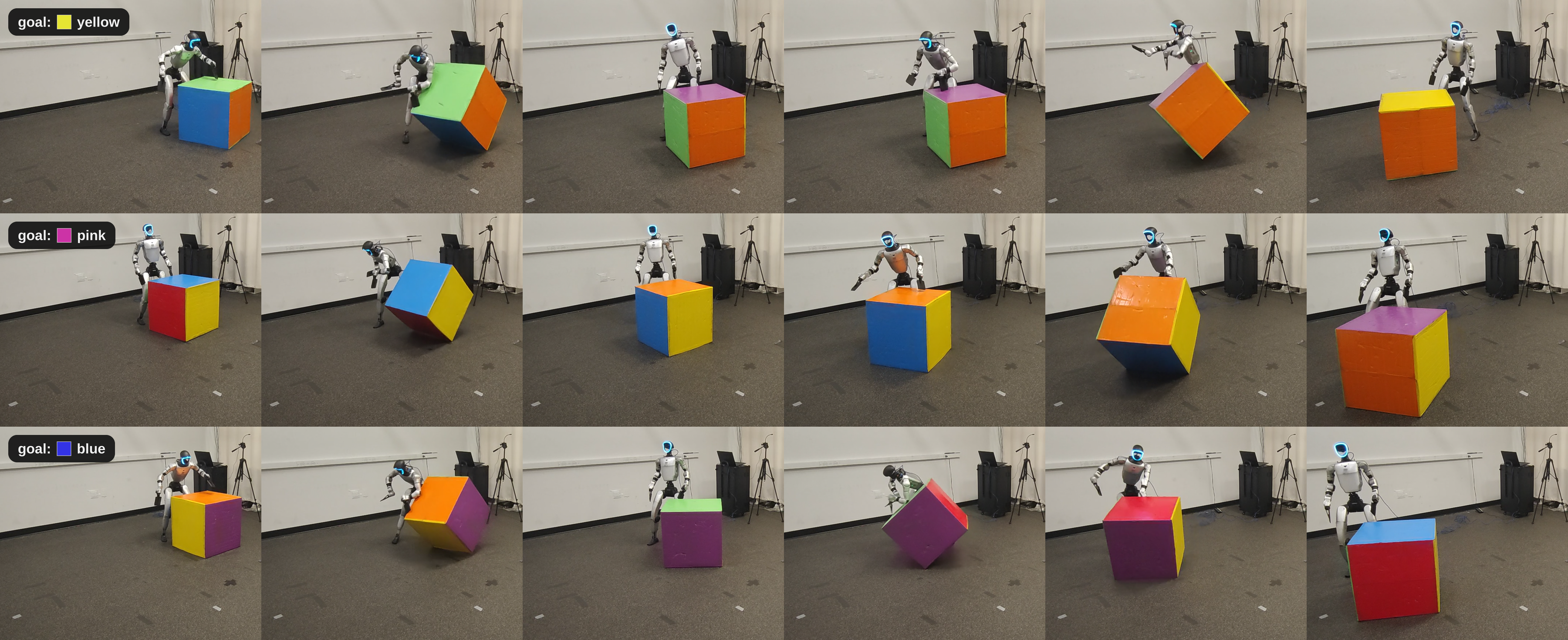}
\caption{\textbf{A simple planner solves repose task} For yellow, pink, and
blue goals (top to bottom), the planner selects fixed routine of reference clips while 
ViBe-adapted perceptive policies handle the rest. Each row shows
one uninterrupted hardware rollout.}
\label{fig:repose-planner}
\end{figure}

\subsection{Ablations}

We use the Repose task for controlled design ablations due to its well-defined success criterion:
the requested face must lie within said threshold ($17.2^\circ$) of vertical.

\begin{figure}[H]
\centering
\includegraphics[width=\textwidth]{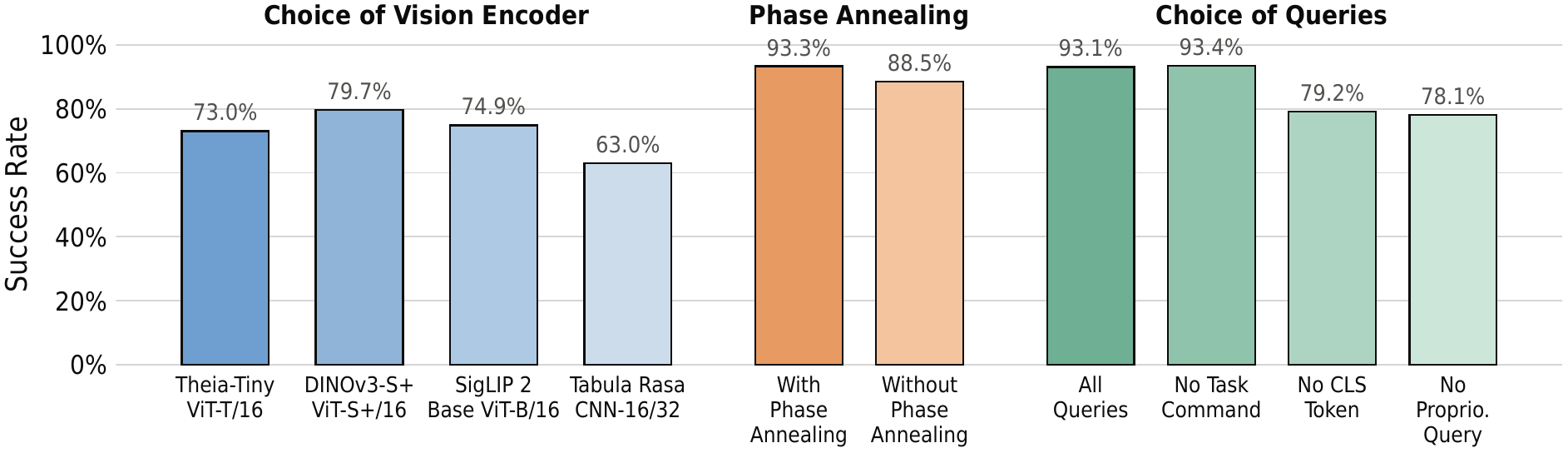}
\caption{\textbf{Controlled ablations on Repose Cube.}
Success rates vs choice of vision encoders (left), with and without phase annealing
(center) and extractor queries (right) }
\label{fig:repose-design-choices}
\end{figure}

\paragraph{Choice of vision encoder.}

To validate the flexibility of our approach, we ablate the choice of vision encoder.
We compare the default Theia-Tiny~\cite{shang2024theia} with 
DINOv3-S+~\cite{simeoni2025dinov3}, 
SigLIP2-B~\cite{tschannen2025siglip2multilingualvisionlanguage}, and a 
CNN trained from scratch~\cite{2946645.2946684}. After an equal number of policy updates,
DINOv3-S+ achieves the highest success rate, followed by SigLIP2-B, Theia-Tiny, and the
tabula-rasa CNN as shown in Fig.~\ref{fig:repose-design-choices} (left). Success across the pre-trained backbones is tightly clustered, with a standard
deviation of only $2.8\%$, while each outperforms the CNN trained from scratch.
Thus, the extractor learns effective visual feedback across backbone choices.

\paragraph{Choice of queries.}

To ablate the contribution of each query, we remove one query at a time and compare the resulting success rates. 
With all three query rows, the policy achieves a success rate of $93.1\%$ at $50k$ updates. Removing the
CLS token or robot proprioception significantly reduces success by $15\%$, showing the significance
of global context and robot's state in extracting task-relevant visual feedback.  Interestingly,
removing the task command query marginally increases success to $93.4\%$, which may
be due to the nature of the chosen task-command for Repose task, demanding further investigation 
in the future across other tasks. Thus, the extractor benefits from the multi-query design. 

\paragraph{Effect of phase annealing.}

Training with phase annealing marginally improves success rate from $88.5\%$ to $93.3\%$ 
as shown in Fig.~\ref{fig:repose-design-choices} (right). However, qualitatively, we found
the curriculum helps keep the learning dynamics stable across different tasks with
varying dynamics and reward landscapes.

\section{Conclusion and Limitations}
\label{sec:conclusion}

We presented ViBe, a post-training framework for visual behavior adaptation of humanoid whole-body
trackers by learning to extract task-relevant perceptive feedback from pre-trained vision
encoders. Equipping a ``blind'' tracker with attention-based perception through low-rank adapters enables
parameter-efficient RL fine-tuning, yielding visual policies
that transfer zero-shot from simulation to hardware. We validate the same recipe across four
tasks spanning non-flat terrain traversal, omni-object loco-manipulation, dodgeball, and
Repose Cube. Together, these experiments demonstrate that perceptive whole-body control can
be achieved by learning only the intermediate connection from scratch while retaining the
pre-trained modules, providing a general recipe for perceptive humanoid control.

In its current form, we note two limitations of our work. First, while the resulting policies are 
visually robust across a wide range of lighting conditions and surface appearances, 
dynamic distractors can still trigger false responses. In dodgeball, for example, the policy 
sometimes mistakes the thrower's head for the ball and dodges unnecessarily. Future work will
target learning necessary invariances through additional domain-randomization events. 
Second, the planner and controller are not
trained in a closed loop. The planner may therefore select a reference whose transitions fall
outside the adapter's training distribution. Although retaining the frozen tracker preserves
reasonable tracking, task-optimal performance would benefit from training with a closed-loop
reference generator~\cite{11246602,zhang2026learningwholebodyhumanoidlocomotion}.

\acknowledgments{We thank Junchao Ma for helping with the hardware enhancements and Shravan
Shenoy for early contributions to this work.}


\bibliography{references}  

\end{document}